\documentclass[sigconf, natbib=false, anonymous=false]{acmart}
\AtBeginDocument{%
  }

\copyrightyear{2025}
\acmYear{2025}
\setcopyright{cc}
\setcctype{by}
\acmConference[MA-LLM '25]{Proceedings of the ACM MM 2025 Workshop on Multimedia Analytics with Multimodal Large Language Models}{October 27--28, 2025}{Dublin, Ireland}
\acmBooktitle{Proceedings of the ACM MM 2025 Workshop on Multimedia Analytics with Multimodal Large Language Models (MA-LLM '25), October 27--28, 2025, Dublin, Ireland}\acmDOI{10.1145/3746263.3757710}
\acmISBN{979-8-4007-2045-1/2025/10}

\RequirePackage[
  datamodel=acmdatamodel,
  style=acmnumeric,
  ]{biblatex}

\usepackage{multirow}

\begin{document}

\title{TriPSS: A Tri-Modal Keyframe Extraction Framework Using Perceptual, Structural, and Semantic Representations}


\author{Mert Can Cakmak}
\affiliation{%
  \institution{Computer and Information Science, University of Arkansas - Little Rock}
  \city{Little Rock}
  \state{Arkansas}
  \country{USA}
}
\email{mccakmak@ualr.edu}

\author{Nitin Agarwal}
\affiliation{%
  \institution{ICSI, University of California, Berkeley}
  \city{Berkeley}
  \state{California}
  \country{USA}
}
\affiliation{%
  \institution{COSMOS Research Center, University of Arkansas - Little Rock}
  \city{Little Rock}
  \state{Arkansas}
  \country{USA}
}
\email{nxagarwal@ualr.edu}

\author{Diwash Poudel}
\affiliation{%
  \institution{Information Science, University of Arkansas - Little Rock}
  \city{Little Rock}
  \state{Arkansas}
  \country{USA}
}
\email{dpoudel@ualr.edu}



\renewcommand{\shortauthors}{Cakmak et al.}

\begin{abstract}
Efficient keyframe extraction is critical for video summarization and retrieval, yet capturing the full semantic and visual richness of video content remains challenging. We introduce TriPSS, a tri-modal framework that integrates perceptual features from the CIELAB color space, structural embeddings from ResNet-50, and semantic context from frame-level captions generated by LLaMA-3.2-11B-Vision-Instruct. These modalities are fused using principal component analysis to form compact multi-modal embeddings, enabling adaptive video segmentation via HDBSCAN clustering. A refinement stage incorporating quality assessment and duplicate filtering ensures the final keyframe set is both concise and semantically diverse. Evaluations on the TVSum20 and SumMe benchmarks show that TriPSS achieves state-of-the-art performance, significantly outperforming both unimodal and prior multimodal approaches. These results highlight TriPSS’s ability to capture complementary visual and semantic cues, establishing it as an effective solution for video summarization, retrieval, and large-scale multimedia understanding.
\end{abstract}

\ccsdesc[500]{Computing methodologies~Video summarization}
\ccsdesc[500]{Information systems~Multimedia and multimodal retrieval}

\keywords{Multimodal Keyframe Extraction,
Video Summarization and Retrieval,
Large Language Models for Video Understanding,
Adaptive Clustering and Visual Analytics,
Multimodal Representation Learning}


\maketitle

\section{Introduction}

The rapid expansion of video platforms, driven by advanced storage, high-speed internet, and widespread mobile adoption, has led to a surge in both short-form and long-format video consumption \cite{zhao2021consumption,violot2024shorts}. This growth of video content offers valuable opportunities for tasks such as summarization, sentiment analysis, and topic extraction, yet it also poses significant computational and storage challenges when entire videos are processed in detail.

Recent approaches incorporate large language models to enhance video understanding. For instance, video models such as VideoLLM \cite{chen2023videollm} and Video-LLaVA \cite{lin2023video} exploit rich representations within video content, yet they remain inefficient when every frame is analyzed. Keyframe extraction addresses this inefficiency by selecting a set of frames that preserve essential information, enabling tasks such as browsing, indexing, and retrieval with reduced redundancy \cite{asha2018key}.

\begin{figure*}[ht]
  \centering
  \includegraphics[width=0.90\linewidth]{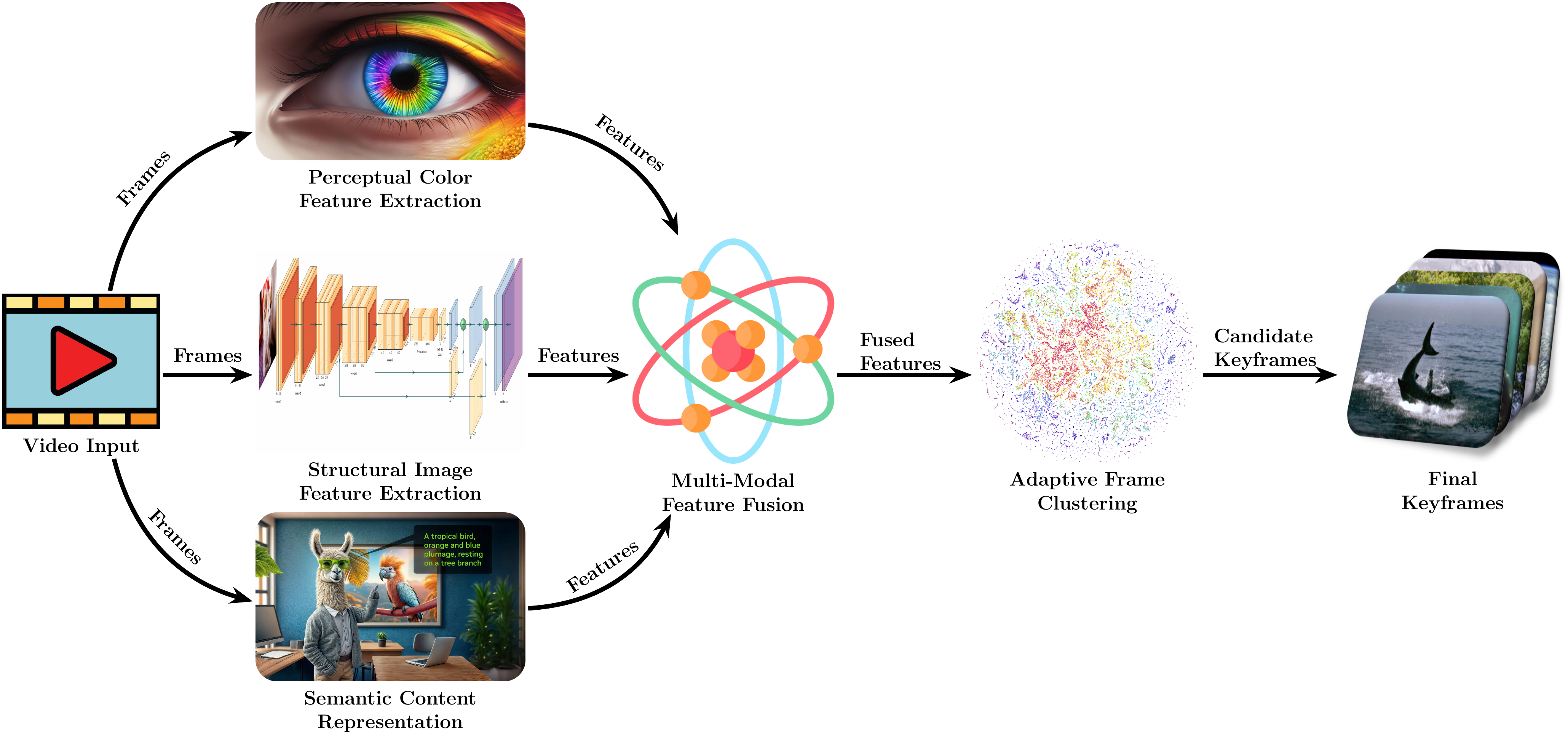}
  \caption{Overview of the TriPSS Framework for Keyframe Extraction}
  \label{diagram}
  \Description{Overview of the TriPSS Framework for Keyframe Extraction}
\end{figure*}

In this work, we present \textbf{TriPSS}, a unified, tri-modal keyframe extraction framework that integrates color \emph{(perceptual)}, CNN-based \emph{(structural)}, and LLM-generated \emph{(semantic)} features to offer a more holistic representation of video content. By harnessing human color perception in the CIELAB space, deep image features extracted via ResNet-50, and frame-level captions generated by a vision-aware model such as Llama Vision, TriPSS captures nuanced semantic and visual details that purely visual methods often overlook, resulting in more precise and interpretable keyframe selection. This stands in contrast to traditional methods that rely on simpler textual or visual cues. Moreover, comprehensive evaluations on benchmark datasets TVSum20 and SumMe demonstrate that TriPSS outperforms state-of-the-art approaches, establishing a new benchmark for keyframe extraction. An overview of the TriPSS process is illustrated in Figure~\ref{diagram}. The implementation of TriPSS is available at \href{https://github.com/Mccakmak/tri-modal-keyframe-extraction}{\textcolor{blue}{GitHub link}}.

\section{Related Work} \label{lit-review}

Keyframe extraction has evolved from basic methods using color histograms and simple clustering \cite{sunuwar2024comparative, asha2018key}, which primarily relied on detecting abrupt visual changes but struggled with complex scenes and lacked semantic depth. To improve flexibility, adaptive clustering techniques \cite{zhao2019key, man2022interested} were introduced, dynamically adjusting cluster boundaries; however, they remain limited by their dependence on unimodal features such as color histograms.

Color-based techniques emphasize perceptual uniformity \cite{asim2018key} to ensure aesthetic coherence, yet they often neglect critical structural and semantic aspects of video content. Human-centered methods that incorporate cognitive signals \cite{bezugam2021efficient} have shown promise, yet they suffer from scalability issues due to reliance on manual annotations. Image feature-based methods focus on structural properties like edges and textures \cite{di2003image}, capturing visual saliency without fully understanding the underlying semantic context.

The integration of large language models (LLMs) and deep learning has further advanced keyframe extraction. For instance, approaches such as those in \cite{park2024too} and \cite{wang2024videotree} focus on enhancing semantic relevance but tend to overlook low-level visual cues. Similarly, models like those proposed in \cite{huang2019novel}, \cite{liang2024keyvideollm}, and \cite{tan2024koala} improve semantic consistency, yet they often miss structural coherence or introduce redundancy. In essence, while these methods excel in semantic understanding, they struggle to provide a comprehensive visual representation.

Keyframe selection techniques also vary in their effectiveness. Histogram-based methods \cite{rodriguez2018selection, de2011vsumm} detect visual changes but miss finer semantic nuances. Structural approaches, such as edge-LBP \cite{nandini2022shot}, perform well in static scenes yet falter in dynamic content. Deep learning models \cite{tang2023deep, tan2024large} enhance temporal analysis but often fail to capture key semantic frames and lack cross-domain generalization. Traditional techniques like k-means \cite{muhammad2020k} offer efficiency for fast-moving videos but remain inflexible.

TriPSS addresses these limitations by integrating perceptual, structural, and semantic representations into a unified framework. Unlike traditional methods that focus on isolated features, TriPSS simultaneously captures color information, deep structural cues, and nuanced semantic context, resulting in enhanced keyframe diversity and representativeness. This comprehensive fusion directly tackles the scalability and adaptability issues inherent in previous approaches, paving the way for more robust video summarization and retrieval.

\section{Methodology} \label{methodology}

In this section, we present a comprehensive multi-modal keyframe extraction pipeline that integrates perceptual color, structural image, and semantic content features through robust feature fusion, density-based clustering, and rigorous candidate keyframe refinement to generate a concise and representative video summary.

\subsection{Perceptual Color Feature Extraction}

To align with human perception, we convert video frames from RGB to the CIELAB (Lab) color space, which offers perceptual uniformity and ensures that numerical color differences correspond closely to those perceived by humans \cite{maji2022perceptually}. We extract color features by computing normalized histograms and the first three statistical moments: mean (intensity), variance (contrast), and skewness (asymmetry) from the L (lightness), a (green to red), and b (blue to yellow) channels \cite{kusumaningrum2014cielab}. Additionally, we compute a colorfulness metric using the mean and standard deviation of the a and b channels to capture visual vibrancy. This compact and perceptually grounded feature set improves clustering quality, enabling more meaningful keyframe selection for summarization.

\subsection{Structural Image Feature Extraction}

In the second phase of our feature extraction process, we use the ResNet-50 v1.5 architecture \cite{he2016deep}, a convolutional neural network known for its strong ImageNet performance \cite{deng2009imagenet} and robust generalization across diverse visual tasks \cite{wightman2021resnet, wilkerson2024benchmarking}. ResNet-50 incorporates residual connections to mitigate vanishing gradients and extract high-level semantic features. We adopt pre-trained weights from the IMAGENET1K\_V2 variant, trained extensively on ImageNet, and remove the final classification layer to extract 2048-dimensional embeddings from the last convolutional block. These structural features complement the perceptual color representations, forming a rich embedding space that supports effective keyframe extraction.

\subsection{Semantic Content Representation Feature Extraction}

Semantic content representation was extracted using the LLaMA-3.2-11B-Vision-Instruct model \cite{Meta2025Llama3}, selected for its strong multimodal capabilities and image understanding performance. We generated frame-level captions using a fixed prompt (“In one sentence, describe the visible content of this provided image”) and deterministic decoding, disabling sampling to ensure reproducible and consistent outputs \cite{zarriess2021decoding}. To handle empty or irrelevant responses, we filtered using predefined keywords and replaced them with “No visible content.” Caption quality was assessed on the COCO dataset \cite{lin2014microsoft} using the CLAIR metric \cite{chan2023clair}; while Phi-3-Vision \cite{microsoft2025phi3vision} slightly outperformed LLaMA (0.72 vs. 0.71), qualitative analysis showed LLaMA produced more semantically aligned descriptions, justifying its use.

We encoded the captions using the all-mpnet-base-v2 model from SentenceTransformers \cite{huggingface-mpnet}, yielding 768-dimensional embeddings. MPNet has shown superior performance on semantic benchmarks such as GLUE \cite{wang2019gluemultitaskbenchmarkanalysis} and SQuAD \cite{rajpurkar2016squad100000questionsmachine}, and captures fine-grained relationships essential for clustering and summarization tasks \cite{song2020mpnet}.

\subsection{Multi-Modal Feature Fusion}
We build a unified frame representation by combining three complementary views: perceptual/color (CIELAB histograms and moments), structural (ResNet-50 embeddings), and semantic (caption embeddings). To balance their scales, each modality is z-score normalized before fusion \cite{kammoun2024impact}. Let the raw vectors be \(f_c\!\in\!\mathbb{R}^{778}\), \(f_i\!\in\!\mathbb{R}^{2048}\), and \(f_t\!\in\!\mathbb{R}^{768}\). After normalization,
\[
f=\hat{f}_c \oplus \hat{f}_i \oplus \hat{f}_t \in \mathbb{R}^{3594},
\]
which serves as our multi-modal feature prior to reduction \cite{cai2025multimodal}.

To obtain a compact and stable space for clustering, we compared several dimensionality reduction options: PCA, random projection, and truncated feature selection. We also probed nonlinear methods (UMAP, t-SNE), but found them unsuitable at this scale due to runtime, sensitivity to hyperparameters, and limited downstream consistency. Across TVSum20 and SumMe, PCA with 512 components provided the best accuracy–efficiency balance; PCA-256 degraded performance, while PCA-1024 offered no clear gains but higher cost. This choice aligns with common practice in multimodal pipelines (e.g., CLIP uses moderate-width projections) \cite{radford2021learning}.

Accordingly, we project with \(W\in\mathbb{R}^{512\times 3594}\) and obtain the reduced representation
\[
f' = W f \in \mathbb{R}^{512},
\]
which we use for adaptive clustering and keyframe selection.

\subsection{Adaptive Frame Clustering}

Clustering was performed using HDBSCAN \cite{campello2013density}, a density-based algorithm well suited for videos with variable scene dynamics. Unlike DBSCAN \cite{ester1996density}, HDBSCAN adapts to varying densities and requires no predefined number of clusters \cite{man2022interested}. Compared to methods like K-Means or Gaussian Mixture Models, it handles irregular cluster shapes and automatically labels transitional or low-quality frames as noise, improving keyframe quality. We evaluated clustering performance using the Density-Based Clustering Validation (DBCV) index \cite{moulavi2014density}, which assesses cohesion and separation based on local density and mutual reachability. A grid search was conducted to tune HDBSCAN hyperparameters for optimal DBCV score while maintaining meaningful scene boundaries.

Formally, given the set of fused and projected frame embeddings \( \{f_1, f_2, \dots, f_N\} \in \mathbb{R}^{512} \), HDBSCAN produces a set of clusters \( \{C_1, C_2, \dots, C_K\} \), where each \( C_k \subseteq \{1, \dots, N\} \). For each cluster \( C_k \), we identify the medoid frame index \( j_k \) as:

\[
j_k = \arg\min_{i \in C_k} \sum_{j \in C_k} \lVert f_i - f_j \rVert_2
\]

The final keyframe set is then defined as \( \mathcal{K} = \{j_1, j_2, \dots, j_K\} \), ordered by frame index to preserve temporal coherence. This medoid-based approach ensures that selected keyframes are real, representative frames from the video rather than synthetic centroids, resulting in concise, diverse, and semantically coherent summaries.

\begin{figure*}[ht]
  \centering
  \includegraphics[width=0.85\linewidth]{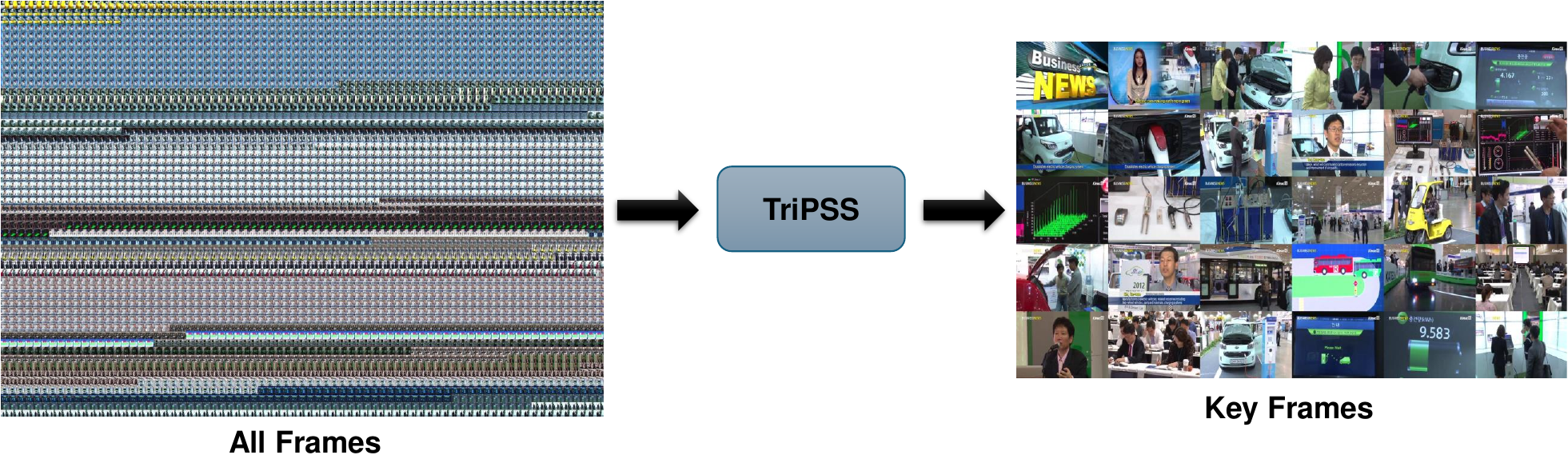}
  \captionsetup{justification=centering} 
  \caption{Example of Keyframe Extraction Demonstrating Temporal Distribution for the Video 
  ``Electric Cars Making Earth More Green'' (video\_id: akI8YFjEmUw) from the TvSum Dataset.}
  \label{keyframe_reduction}
  \Description{Example of Keyframe Extraction Demonstrating Temporal Distribution for the Video}
\end{figure*}

\subsection{Refined Keyframe Selection}

Candidate frames were refined through quality assessment and duplicate filtering. Extreme low-light frames were discarded based on grayscale intensity, variance, and structural integrity using the Canny edge detector \cite{amer2015edge} and Laplacian variance. Color uniformity was measured via histogram variance, and visual saliency was assessed by comparing central and global intensity. Since text often conveys crucial contextual information, especially in frames with minimal visual details, text presence was detected using the MSER algorithm \cite{icsik2014comparative} and verified with ORB keypoint detection \cite{sharma2023comparative}.

Duplicate filtering employed the Structural Similarity Index Measure (SSIM) \cite{bakurov2022structural} to remove redundant frames by considering luminance, contrast, and structure. We relied on these traditional computer vision techniques for their computational efficiency and proven robustness, which complement HDBSCAN's noise handling and clustering capabilities. This integrated approach effectively filtered out low-quality frames while preserving those that offered a comprehensive visual summary, ensuring fast and reliable keyframe extraction for video summarization and retrieval.

\section{Evaluation}\label{evaluation}

We assess TriPSS on two established benchmarks: TVSum20 \cite{tan2024large} and SumMe \cite{gygli2014creating}. TVSum20 consists of 20 consumer-grade videos annotated with 1,000 shot-level importance scores, while SumMe contains 25 user-generated videos with importance ratings from at least 15 human annotators. Both datasets are widely used in video summarization and enable fair, keyframe-based evaluation. Figure~\ref{keyframe_reduction} visualizes how TriPSS identifies a concise and representative subset of frames, effectively filtering redundancy while preserving semantic richness.

We adopt F1 with frame-level similarity: a ground-truth keyframe \(h_i \in \mathcal{K}^*\) and a predicted keyframe \(h_j \in \hat{\mathcal{K}}\) match if the cosine similarity \( \mathrm{sim}(h_i,h_j) = \frac{h_i^\top h_j}{\lVert h_i\rVert \lVert h_j\rVert} \) is at least \( \tau = 0.9 \). Let \(M\) be the set of matched pairs; we report \( \mathrm{F1} = \frac{2|M|}{|\mathcal{K}^*| + |\hat{\mathcal{K}}|} \). This evaluates keyframes by perceptual similarity rather than index overlap.

Table~\ref{tab:ablation} presents the results of an ablation study analyzing the contribution of each modality. Removing any single component which are perceptual (Pe), structural (St), or semantic (Se) leads to a clear drop in F1-score, confirming that each modality contributes complementary information. Structural and semantic cues individually outperform the perceptual stream, while combinations of two modalities improve results further. The full tri-modal configuration consistently achieves the best performance, with F1 scores of 0.6104 on TVSum20 and 0.5902 on SumMe, validating the effectiveness of multi-modal integration in keyframe selection.

\begin{table}[ht]
  \caption{Ablation study: modality contributions (F1) on TVSum20 and SumMe.}
  \label{tab:ablation}
  \begin{tabular}{lc|lc}
    \toprule
    \multicolumn{2}{c}{TVSum20} &
    \multicolumn{2}{c}{SumMe} \\
    \cmidrule(lr){1-2} \cmidrule(lr){3-4}
    Variant & F1 &
    Variant & F1 \\
    \midrule
    Pe (Perceptual only)      & 0.4382 & Pe (Perceptual only)      & 0.3794 \\
    Se (Semantic only)        & 0.4867 & Se (Semantic only)        & 0.4295 \\
    St (Structural only)      & 0.5114 & St (Structural only)      & 0.4612 \\
    Pe + Se                   & 0.5458 & Pe + Se                   & 0.4973 \\
    Pe + St                   & 0.5590 & Pe + St                   & 0.5107 \\
    St + Se                   & 0.5843 & St + Se                   & 0.5388 \\
    \midrule
    \textbf{Pe + St + Se (TriPSS)} & \textbf{0.6104} &
    \textbf{Pe + St + Se (TriPSS)} & \textbf{0.5902} \\
    \bottomrule
  \end{tabular}
\end{table}

We also compare TriPSS to a range of established baselines in Table~\ref{results}. These include classical statistical methods, learning-based summarization approaches, and recent transformer-driven or multi-modal models. TriPSS outperforms all competitors across both datasets, with particularly strong gains over unimodal techniques. This improvement highlights the advantage of fusing perceptual, structural, and semantic signals, as well as the impact of our adaptive clustering and post-processing strategies in eliminating redundancy while retaining content diversity.

\begin{table}[ht]
  \caption{Performance Comparison of Keyframe Extraction Methods on TvSum20 and SumMe Datasets}
  \label{results}
  \begin{tabular}{lc|lc}
    \toprule
    \multicolumn{2}{c}{\textbf{TvSum20}} & \multicolumn{2}{c}{\textbf{SumMe}} \\
    \cmidrule(lr){1-2} \cmidrule(lr){3-4}
    \textbf{Method} & \textbf{F1} & \textbf{Method} & \textbf{F1} \\
    \midrule
    HistDiff \cite{rodriguez2018selection}  & 0.3380 & H-MAN \cite{liu2019learning}        & 0.5180 \\
    VS-UID \cite{garcia2023videosum}         & 0.4615 & SUM-GDA \cite{li2021exploring}     & 0.5280 \\
    GMC \cite{gharbi2017key}                 & 0.4833 & STVS \cite{kashid2024stvs}         & 0.5360 \\
    VSUMM \cite{de2011vsumm}                 & 0.4894 & TAC-SUM \cite{huynh2023cluster}    & 0.5448 \\
    KMKey  \cite{muhammad2020k}              & 0.5039 & PGL-SUM \cite{apostolidis2021combining} & 0.5560 \\
    LBP-Shot \cite{nandini2022shot}          & 0.5050 & SMN \cite{wang2019stacked}         & 0.5830 \\
    VS-Inception \cite{garcia2023videosum}   & 0.5168 & AugFusion \cite{psallidas2023video} & 0.5840 \\
    LMSKE \cite{tan2024large}                & 0.5311 & Ldpp-c \cite{kaseris2022exploiting} & 0.5880 \\
    \midrule
    \textbf{TriPSS}                          & \textbf{0.6104} & \textbf{TriPSS} & \textbf{0.5902} \\
    \bottomrule
  \end{tabular}
\end{table}

\section{Conclusion} \label{conclusion}

We presented TriPSS, a tri-modal keyframe extraction framework that integrates perceptual features (CIELAB), structural embeddings (ResNet-50), and semantic representations (LLaMA-3.2-11B-Vision-Instruct). Using z-score normalization, PCA-based fusion, and adaptive clustering with HDBSCAN, TriPSS generates compact, semantically rich video summaries, achieving state-of-the-art results on TVSum20 and SumMe. While effective, it currently relies on simple feature concatenation and lacks temporal modeling. Future work will explore attention-based fusion, sequence-aware summarization, and interactive frameworks. Beyond summarization, TriPSS demonstrates how integrating vision and language enables scalable, interpretable multimedia analytics. Its modular design and compatibility with large multimodal models make it well-suited for interactive systems and human-in-the-loop applications, aligning closely with the goals of next-generation multimedia analysis.

\begin{acks}

This research is funded in part by the U.S. National Science Foundation (OIA-1946391, OIA-1920920), U.S. Office of the Under Secretary of Defense for Research and Engineering (FA9550-22-1-0332), U.S. Army Research Office (W911NF-23-1-0011, W911NF-24-1-0078, W911NF-25-1-0147), U.S. Office of Naval Research (N00014-21-1-2121, N00014-21-1-2765, N00014-22-1-2318), U.S. Air Force Research Laboratory, DARPA, the Australian DoD Strategic Policy Grants Program, Arkansas Research Alliance, the Jerry L. Maulden/Entergy Endowment, and the Donaghey Foundation at UA Little Rock. Opinions are the authors’ own and do not necessarily reflect the funders; we gratefully acknowledge their support.
\end{acks}

\printbibliography


\end{document}